\documentclass[letterpaper, 10 pt, conference]{ieeeconf}  % Comment this line out if you need a4paper

\IEEEoverridecommandlockouts                              % This command is only needed if 
\usepackage{graphics}
\usepackage{multirow}

\usepackage{graphics} % for pdf, bitmapped graphics files
\usepackage{epsfig} % for postscript graphics files
\usepackage{times} % assumes new font selection scheme installed
\usepackage{amsmath} % assumes amsmath package installed
\usepackage{amssymb}  % assumes amsmath package installed

\usepackage{array}
\usepackage{booktabs}
\usepackage{siunitx}

\usepackage[hidelinks]{hyperref}
\usepackage[T1]{fontenc}  % To make copy of underscore work

\usepackage{xcolor,pifont}
\usepackage{color}
\usepackage{acronym}

\title{\LARGE \bf

High-Frequency Capacitive Sensing for Electrohydraulic Soft Actuators}

\author{
 Michel R. Vogt$^{1\dag}$, 
 Maximilian Eberlein$^{1\dag}$,
 Clemens C. Christoph$^{1\dag}$,
 Felix Baumann$^{1}$, 
 Fabrice Bourquin$^{1}$, \\
 Wim Wende$^{1}$,
 Fabio Schaub$^{1}$,
 Amirhossein Kazemipour$^1$, % % supervision
 Robert K. Katzschmann$^{1*}$% % supervision
\thanks{$^{1}$Soft Robotics Lab, ETH Zurich, Switzerland
        {\tt\footnotesize 
        \{\href{mailto:micvogt@ethz.ch}{micvogt},
        \href{mailto:meberlein@ethz.ch}{meberlein},
        \href{mailto:cchristoph@ethz.ch}{cchristoph},
        \href{mailto:febaumann@ethz.ch}{febaumann},
        \href{mailto:fbourquin@ethz.ch}{fbourquin},
        \href{mailto:wwende@ethz.ch}{wwende},
        \href{mailto:faschaub@ethz.ch}{faschaub},
        \href{mailto:akazemi@ethz.ch}{akazemi},
        \href{mailto:rkk@ethz.ch}{rkk}\}@ethz.ch}}%
\thanks{$^{\dag}$Equal Contribution}
\thanks{$^{*}$Corresponding author: \href{mailto:rkk@ethz.ch}{\tt rkk@ethz.ch}}
}

\begin{document}

\maketitle
%%%%%%%%%%%%%%%%%%%%%%%%%%%%%%%%%%%%%%%%%%%%%%%%%%%%%%%%%%%%%%%%%%%%%%%%%%%%%%%%

\begin{abstract}
The need for compliant and proprioceptive actuators has grown more evident in pursuing more adaptable and versatile robotic systems. Hydraulically Amplified Self-Healing Electrostatic (HASEL) actuators offer distinctive advantages with their inherent softness and flexibility, making them promising candidates for various robotic tasks, including delicate interactions with humans and animals, biomimetic locomotion, prosthetics, and exoskeletons. This has resulted in a growing interest in the capacitive self-sensing capabilities of HASEL actuators to create miniature displacement estimation circuitry that does not require external sensors. However, achieving HASEL self-sensing for actuation frequencies above 1 Hz and with miniature high-voltage power supplies has remained limited. In this paper, we introduce the F-HASEL actuator, which adds an additional electrode pair used exclusively for capacitive sensing to a Peano-HASEL actuator. We demonstrate displacement estimation of the F-HASEL during high-frequency actuation up to 20 Hz and during external loading using miniaturized circuitry comprised of low-cost off-the-shelf components and a miniature high-voltage power supply. Finally, we propose a circuitry to estimate the displacement of multiple F-HASELs and demonstrate it in a wearable application to track joint rotations of a virtual reality user in real-time.

\end{abstract}
%%%%%%%%%%%%%%%%%%%%%%%%%%%%%%%%%%%%%%%%%%%%%%%%%%%%%%%%%%%%%%%%%%%%%%%%%%%%%%%%

\section{Introduction}\label{introduction}

\begin{figure}[t!]
\centering   
\includegraphics[width=\linewidth]{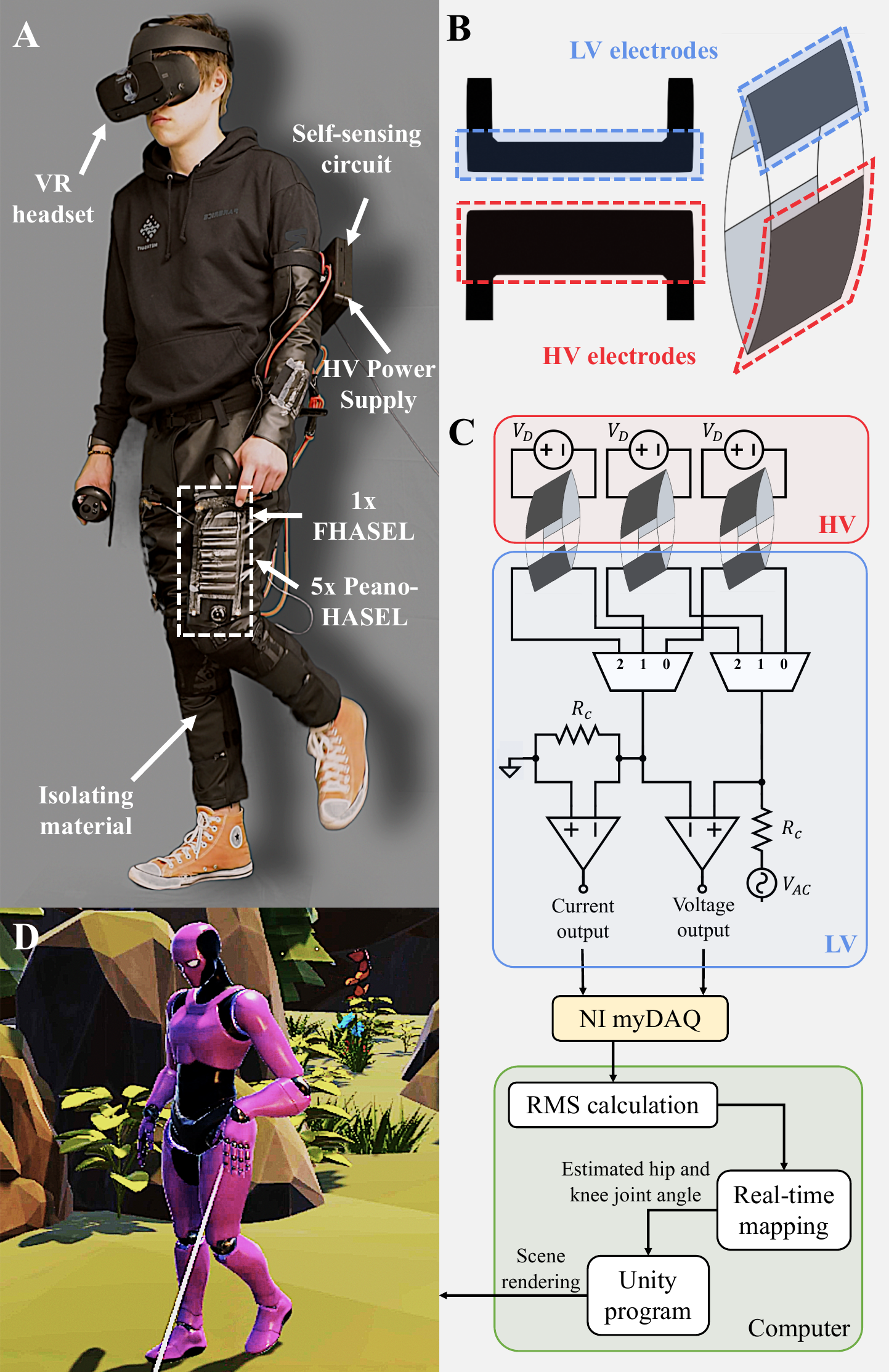}
\caption{(A) Wearable VR application to track the users knee and hip joint rotation in a 2D-plane (B) Proposed F-HASEL, characterized by an additional electrode used exclusively for sensing. (C) Visualization of the methodology behind the wearable VR application, composed of the schematic of the sensing circuitry which enables quasi-simultaneous displacement estimation of multiple F-HASEL actuators as well as the subsequent data processing. (D) Virtual avatar to which the users tracked joint rotations are transmitted.
}
\label{fig:1}
\vspace{-20pt}
\end{figure}

\begin{figure}[t]
    \centering
    \includegraphics[width=0.5\textwidth]{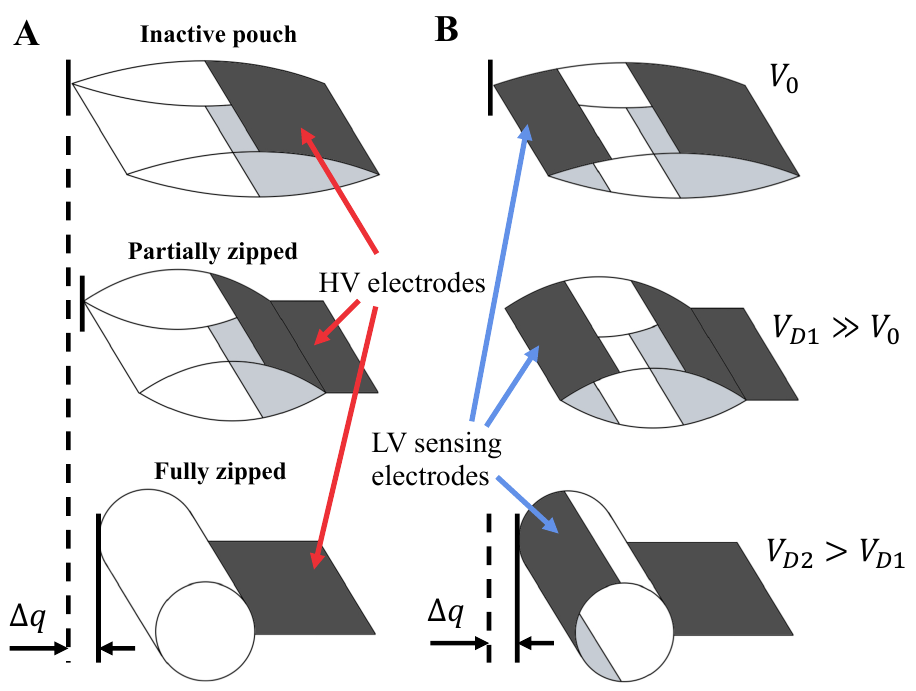}
    \caption{The traditional Peano-HASEL and F-HASEL actuators both undergo distinct stages of actuation with increasing driving voltages, but  F-HASELs also integrate sensing electrodes for improved control and feedback. (A) Traditional Peano-HASEL actuator at different stages of actuation caused by the driving voltages \(V_0\), \(V_{D1}\), and \(V_{D2}\) of increasing magnitude applied to the HV electrodes. (B) F-HASEL actuator, characterized by the LV sensing electrodes, at different stages of actuation caused by the driving voltages \(V_0\), \(V_{D1}\), and \(V_{D2}\) of increasing magnitude applied to the HV electrodes.}
    \label{fig:frog-hasel}
\end{figure}

A commonly used type of HASEL is the Peano-HASEL actuator \cite{HaselPeano,HaselProgress} consisting of a flexible, non-stretchable shell, partially covered in a symmetrical pair of electrodes and filled with a dielectric liquid (Fig. \ref{fig:frog-hasel}A). When a high enough driving voltage is applied to the electrode pair the electrodes zip together, pushing fluid into the remaining portion of the pouch which results in a contraction of the actuator (Fig. \ref{fig:frog-hasel}A). This linear contraction will be referred to as the displacement of the actuator. In HASEL actuators as a whole, a promising technique for estimating the displacement of the actuator is capacitive self-sensing \cite{Hasel,SelfSenseClosedLoop,ly2021miniaturized}. A common technique used for capacitive self-sensing of electrostatic transducers, which has been widely demonstrated for DEAs \cite{Dea1,Dea2,Dea3,Superimpose}, is based on superimposing a low-voltage (LV) sinusoidal sensing signal onto the high-voltage (HV) driving signal applied to the electrodes of the actuator, and measure changes in the LV signal caused by the change in capacitance between the electrodes as the actuator deforms. However, these methods require HV electrical components that are either expensive, bulky, or both.

To address this, Ly et al. \cite{ly2021miniaturized} proposed a miniaturized sensing circuitry for electrostatic transducers in which both the superimposition and measurement of the LV sensing signal occur on the LV side of an electrostatic transducer. This facilitates the use of LV off-the-shelf components, reducing both cost and form factor of the circuitry. Furthermore, Ly et al. demonstrated the capability of this circuitry to estimate the displacement of Peano-HASEL actuators. However, when used with Peano-HASEL actuators this method experiences a phase lag between the true displacement and the estimated displacement at actuation frequencies above 1 Hz, resulting in a drastic loss of sensing performance at higher frequencies. Furthermore, the proposed method shows distortions in the estimated capacitance, and consequently the estimated displacement, when driving voltages above \(\approx\) \SI{3.3}{kV} are applied using a miniature, \SI{8}{cm} long HV power supply (XP Power; EMCO C80). While this level of distortion at that voltage-level was not present using the far larger \SI{1.5}{m} tall Trek 50/12 HV amplifier (Trek, Inc.), such a large HV power supply limits the use of the circuitry in untethered and portable applications.

In an attempt to enable accurate displacement estimation at higher frequencies, Karrer \cite{resistivesensor} integrated a resistive sensor onto a Peano-HASEL which was able to show good congruency with the ground truth displacement for sinusoidal actuation at frequencies up to \SI{10}{Hz}. However, the method requires extra fabrication steps and displacement estimation is not possible without knowing the load applied to the actuator.

\subsection{Contributions}

In this paper, we introduce a new HASEL design, the F-HASEL, which, in contrast to a Peano-HASEL, decouples sensing from actuation using an additional electrode pair on the actuator. We propose two compact capacitive sensing circuit designs using low-cost off-the-shelf components. We demonstrate the capability of the circuits to estimate the displacement during high-frequency actuation with a miniature power supply at frequencies up to 20 Hz, as well as the displacement caused by an eccentric contraction through a change in external tensile forces applied to the actuator. Furthermore, the proposed sensing methods allow for the reversal of the driving voltage polarity after actuation cycles. Finally, we propose a circuitry to estimate the displacement of multiple F-HASEL actuators using minimal additional components and demonstrate its capabilities with a wearable virtual reality (VR) application that utilizes multiple F-HASEL actuators to track the joint rotation of the knee and hip joints of a VR user in a 2D-plane.

\section{Methodology}\label{selfsensing}

\subsection{Separating sensing from actuation}

The electrodes on a Peano-HASEL actuator act as a variable capacitor, the capacitance of which is dependent on the electrodes position relative to each other. As such the capacitance changes as the actuator deforms. Previous work has focused on determining the capacitance of the HV electrodes during actuation to estimate the displacement caused by the actuator deforming \cite{ly2021miniaturized}. The capacitance is calculated through measuring changes in a LV sinusoidal signal applied over the HV electrodes. However, this method experiences a phase lag between the true displacement and the estimated displacement at actuation frequencies above 1 Hz, which the authors argue is caused by the non-linear zipping behaviour of the HV electrodes (Fig. \ref{fig:frog-hasel}A).

To address this, we propose a new HASEL design coined the Frog-HASEL (F-HASEL), in which an additional pair of electrodes is added to a Peano-HASEL, on the opposite side of the electrodes used for high-voltage actuation, as shown in Fig. \ref{fig:1}B. Similar to previous work we estimate the displacement of the HASEL by measuring changes in a LV sinusoidal signal. However, unlike in previous work, the LV signal is applied not across the HV electrodes, but across the second pair of electrodes added, which notably does not undergo any zipping during actuation. This added second pair of electrodes will be referred to as the LV sensing electrodes going forward.

One major benefit this design offers over self-sensing designs that are implemented with a single pair of electrodes, is that in our case the LV sensing electrodes over which the LV sinusoidal signal is applied are at a low voltage-level. This allows us to take a direct measurement of the voltage drop (\(V_H\)) of the sensing signal over the LV sensing electrodes with low-voltage off-the-shelf components. Furthermore, unlike other embedded sensors, there is minimal to no increase in production cost, number of fabrication steps, time, or weight of the actuator. If the electrode is applied to the actuator through e.g. screen printing, the only thing that needs to be adjusted is the stencil used in the procedure.

In addition to measuring the voltage over the LV sensing electrodes, we are able to implement a low-side current sensor. This is achieved by placing a resistor (\(R_C\)) in series with the LV sensing electrodes, and measuring the voltage drop (\(V_C\)) over said resistor. The assumption made is that no leakage current is exchanged between the LV sensing electrodes and the HV electrodes to which the driving voltage is applied. Thus, any current flowing through the current-sensing resistor (\(R_C\)) must have flown through the LV sensing electrodes on the F-HASEL actuator.

We electrically model the LV sensing electrodes as a variable capacitor with capacity (\(C_E\)), formed by the electrodes, and a series resistor with resistance (\(R_E\)), representing the resistance of the electrodes. We do not include a parallel resistor in the model as the leakage current over the LV sensing electrodes is assumed to be negligible. As with a traditional Peano-HASEL actuator, when a high enough driving voltage is applied over the HV electrode pair used for actuation, the HASEL will deform and push the fluid into the rest of the pouch on which the LV sensing electrodes are placed. As a result, the distance between the LV sensing electrodes increases and their variable capacity (\(C_E\)) decreases. This change in capacity (\(C_E\)) leads to changes in the voltage and current measurements we obtain, which can be used to estimate a displacement in a variety of ways. In this paper we will focus on two of those methods, which are (1) the 'voltage-method', estimating the displacement by directly mapping them to a change in voltage measurement (\(V_H\)) caused by a filter response without any additional calculations, and (2) the 'impedance-method' using voltage and current information to calculate changes in the complex impedance of the LV sensing electrodes.

\subsection{The voltage-method. Estimating actuator displacement through changes in the filter response}

Using the chosen electrical model and placing additional resistances, e.g. the current-sensing resistor, with a combined resistance (\(R\)) in series with the LV sensing electrodes, we can create a passive first-order low-pass filter described by its cutoff frequency:

\begin{equation}
f_c = \frac{1}{2 \pi (R + R_E) C_E} \label{eq:myequation1}
\end{equation}

A displacement of the F-HASEL actuator (\(\Delta q\)) causes a change in capacitance (\(C_E\)), which results in a shift of the cutoff frequency (\(f_c\)). This has an effect on the voltage gain of the low-pass filter, reflected in the measured voltage (\(V_H\)). Thus, the measured change in voltage gain is directly proportional to the actuator's change in displacement.

\subsection{The impedance-method. Estimating actuator displacement through changes in the complex impedance}

When a low-voltage sinusoidal signal is applied to the LV sensing electrodes, the relationship between the applied voltage and the resulting current is determined by the impedance of the LV sensing electrodes. Having obtained the voltage measurements (\(V_H\)) and (\(V_C\)) over the LV sensing electrodes and the current resistor (\(R_C\)) respectively, we can calculate the magnitude of the complex impedance:

\begin{equation}
\mid Z \mid = \frac{\mid \text{Voltage} \mid}{\mid \text{Current} \mid} = \frac{V_{Hrms} R_C}{V_{Crms}} \label{eq:impedance}
\end{equation}

where \(V_{Hrms}\) and \(V_{Crms}\) are the calculated root mean square values of the voltage measurements (\(V_H\)) and (\(V_C\)) respectively. Furthermore, the complex impedance is in a direct relationship with the electrode resistance (\(R_E\)), the frequency of the sinusoidal sensing signal (\(f\)), as well as the variable capacity (\(C_E\)):

\begin{equation}
\mid Z \mid  = \sqrt{{R_E}^2 + \frac{1}{{2 \pi f}^2} \frac{1}{{C_E}^2}} \label{eq:impedance_mag}
\end{equation}

Under the assumption that the electrode resistance (\(R_E\)) and the frequency of the sensing signal (\(f\)) are fixed, the change in magnitude of the complex impedance is inversely proportional to the change in capacity (\(C_E\)) and thus directly proportional to the change in displacement (\(\Delta q\)) of the F-HASEL actuator.

\subsection{Complete sensing circuitry}

\begin{figure}[t]
    \centering
    \includegraphics[width=0.5\textwidth]{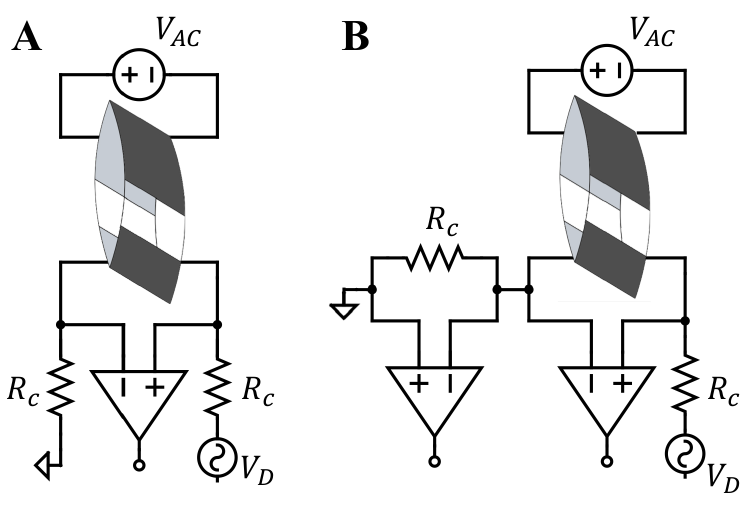}
    \caption{Schematic of the sensing circuitry to enable the displacement estimation of a single F-HASEL actuator using two different methods: (A) the voltage-method, and (B) the impedance-method.}
    \label{fig:circuit}
\end{figure}

\begin{figure}[t]
    \centering
    \includegraphics[width=0.5\textwidth]{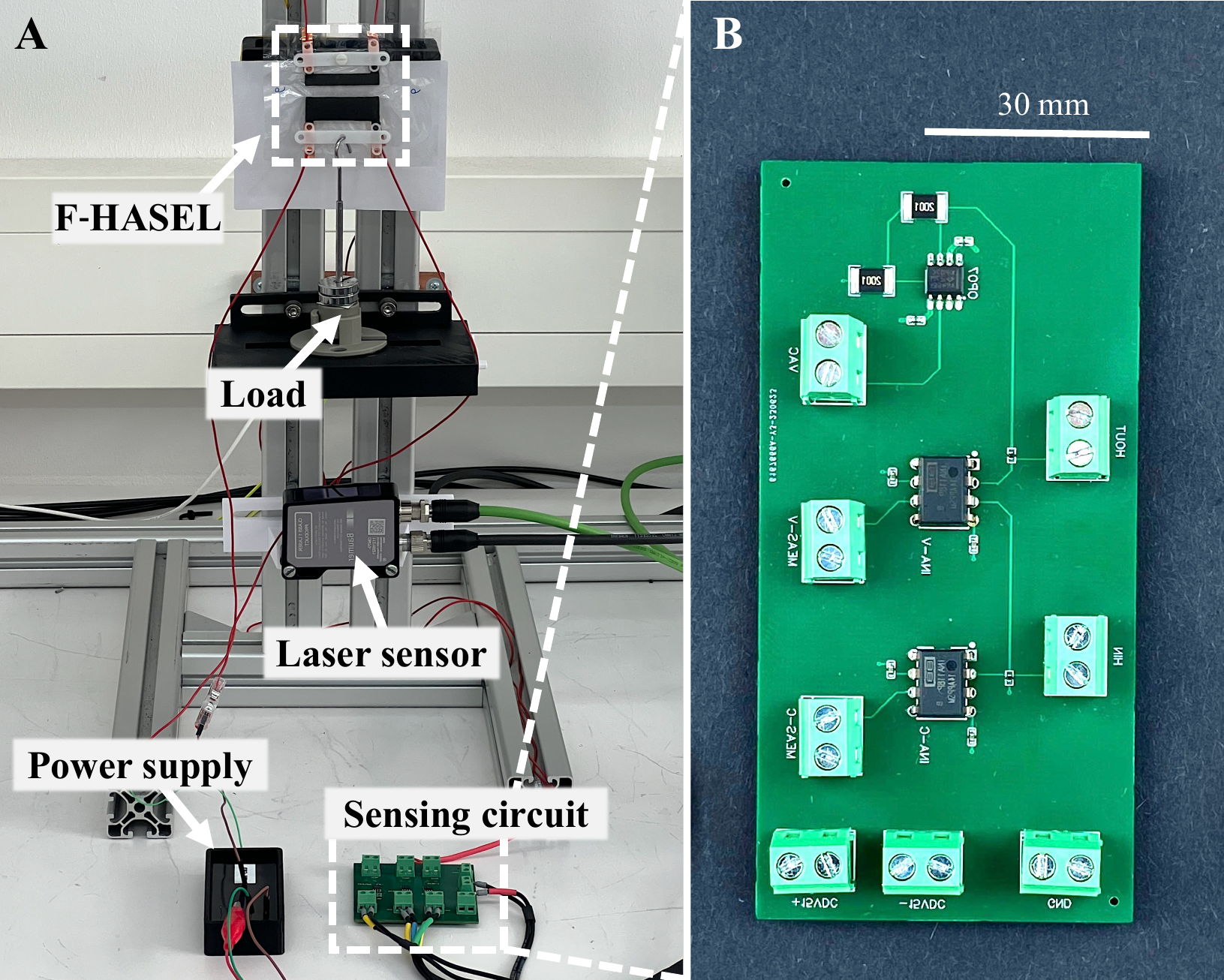}
    \caption{Experimental displacement estimation against ground truth across various actuation cycles, while the sensing circuitry facilitates the estimation for a single actuator. (A) Experimental setup to compare the displacement estimation to the ground truth for a variety of actuation cycles. (B) PCB of the sensing circuitry to enable displacement estimation of a single F-HASEL actuator.}
    \label{fig:setup}
\end{figure}

The circuit takes a LV sinusoidal signal as an input. A non-inverting op amp circuit (OP07CD, Texas Instruments) amplifies the signal with a gain of 1:2. Two instrumentation amplifiers (INA821, Texas Instruments) are used to measure the voltages across the LV sensing electrodes as well as over the current resistor \(R_C\) respectively. Additionally, a second resistor equal in resistance to the current resistor \(R_C\) is placed in series with the LV sensing electrodes on the high side of the sensing signal path. The schematic of this circuitry is shown in Fig. \ref{fig:circuit}B. If current measurements are not required, e.g. if only the voltage-method is used, one of the instrumentation amplifiers can be omitted to simplify the circuitry as shown in Fig. \ref{fig:circuit}A. The small form factor of all of these components contribute to the entire circuitry fitting onto a miniature PCB (Fig. \ref{fig:setup}B).

\subsection{Driving voltage polarity agnostic character of sensing circuitry}

As is shown in Fig. \ref{fig:circuit}A-B, the HV electrodes, to which a HV driving signal is applied, are not in the signal path of our sensing circuit. As such, our proposed sensing methods are driving voltage polarity agnostic. This differentiates them from previous methods like the one proposed by Ly et al. \cite{ly2021miniaturized}, which rely on a specific side of the actuator to stay at low-voltage levels at all times. The ability to reverse the polarity of the driving voltage after actuation cycles is essential for HASEL actuators, as it allows the mitigation of charge retention effects which can prevent the HASEL actuator from fully returning to its initial position \cite{HaselPeano} \cite{chargeretention1}.

\subsection{Circuitry to enable quasi-simultaneous displacement estimation of multiple F-HASEL actuators}

In certain applications, e.g. soft wearables or prosthetic devices, extra importance is placed on the design of a circuitry with a small form factor. This becomes an increasing challenge when multiple HASEL actuators are used in a design, each of which increases the hardware requirements. In a step towards the miniaturization of the sensing circuitry of HASEL actuators, we propose a circuitry that allows the previously presented voltage and current measurements to be performed for multiple F-HASELs using minimal additional components. For this, we switch through an array of different HASELs using two multiplexers (ADG405, Analog Devices), which are placed between the HASELs and the sensing circuitry, as shown in Fig. \ref{fig:1}C. This allows us to use the same measuring components and LV sinusoidal signal generator for all HASELs that we want to estimate the displacement of. This however results in the displacement update rate for an individual HASEL being inversely proportional to the amount of HASELs connected to the circuitry, as the displacement of only one HASEL is being estimated at any given point in time. This design offers a great reduction in the form factor of the required sensing circuitry, but is limited by the update rate requirements of the application.

\begin{figure}[t]
    \centering
    \includegraphics[width=0.5\textwidth]{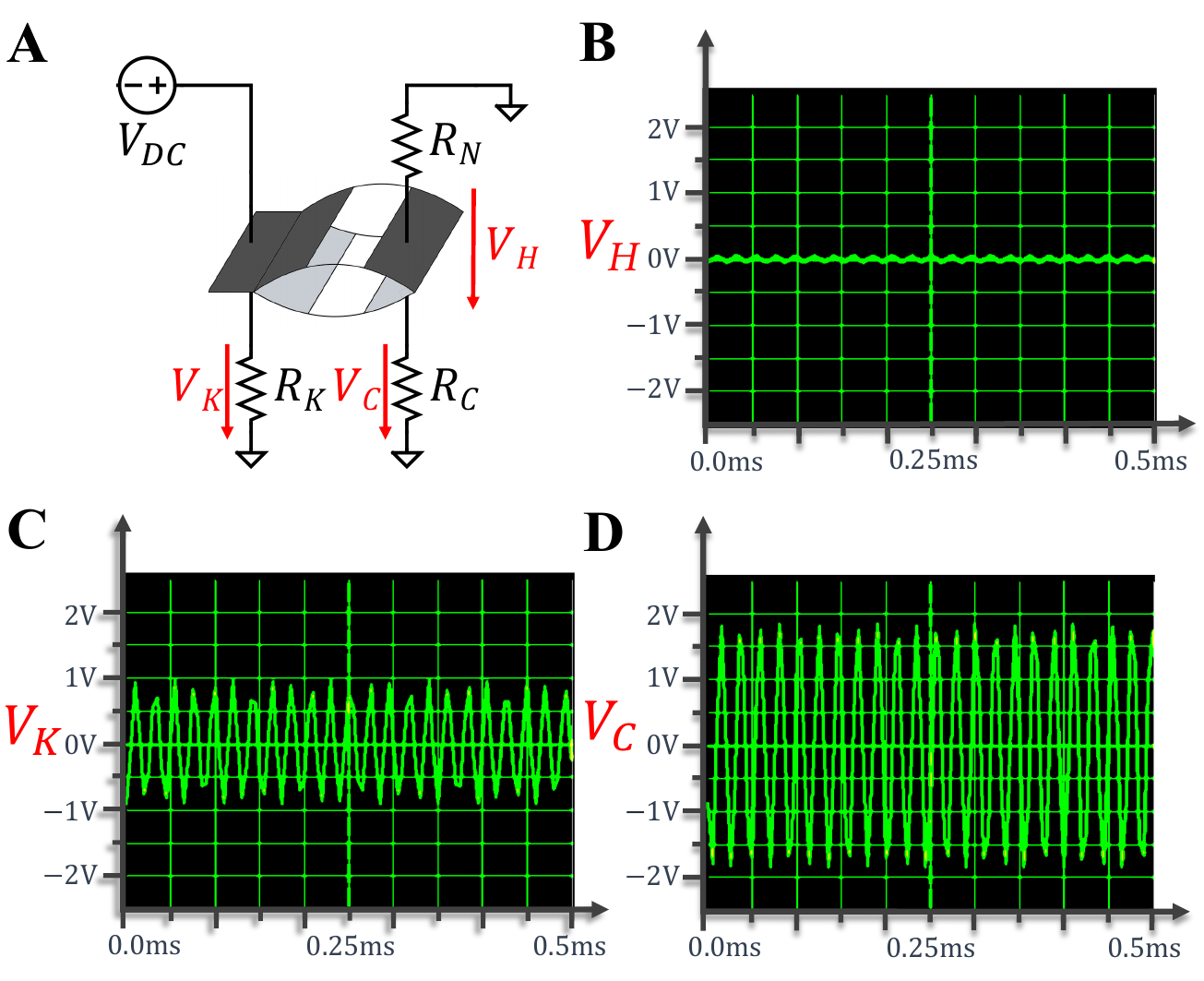}
    \caption{The experiments illustrate noise reduction in differential voltage measurements. (A) Experimental setup to show the difference in noise present on the differential voltage measurements during a constant driving voltage of magnitude \SI{4.8}{kV} in the proposed method compared to a previous work by Ly et al. \cite{ly2021miniaturized}. (B-D) Recorded differential voltage measurements \(V_K\), \(V_C\), and \(V_H\) over a time frame of \SI{0.5}{ms} during a constant driving voltage of magnitude \SI{4.8}{kV}.}
    \label{fig:4}
\end{figure}

\section{Experiments and Results}\label{experiments}
\subsection{Comparing noise magnitude in our measurements with the state-of-the-art}

\begin{figure*}[t]
\centering
\includegraphics[width=\textwidth]{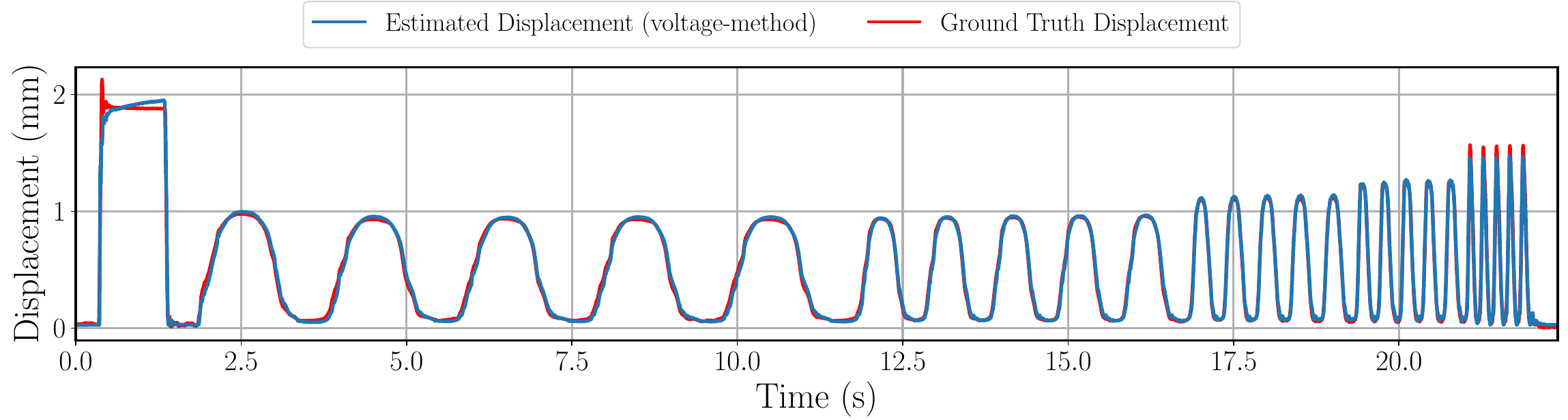}
   \caption{Displacement estimation using the voltage-method during a step actuation with \SI{2}{kV} amplitude and  \SI{2.5}{kV} offset, followed by sinusoidal actuation with amplitude \SI{1}{kV} and offset \SI{3.5}{kV} at frequencies of \SI{0.5}{Hz}, \SI{1}{Hz},\SI{2}{Hz}, \SI{3}{Hz}, and \SI{5}{Hz} respectively.}
   \label{fig:plot1}
\end{figure*}

\begin{figure}[ht]
\centering
\includegraphics[width=0.49\textwidth]{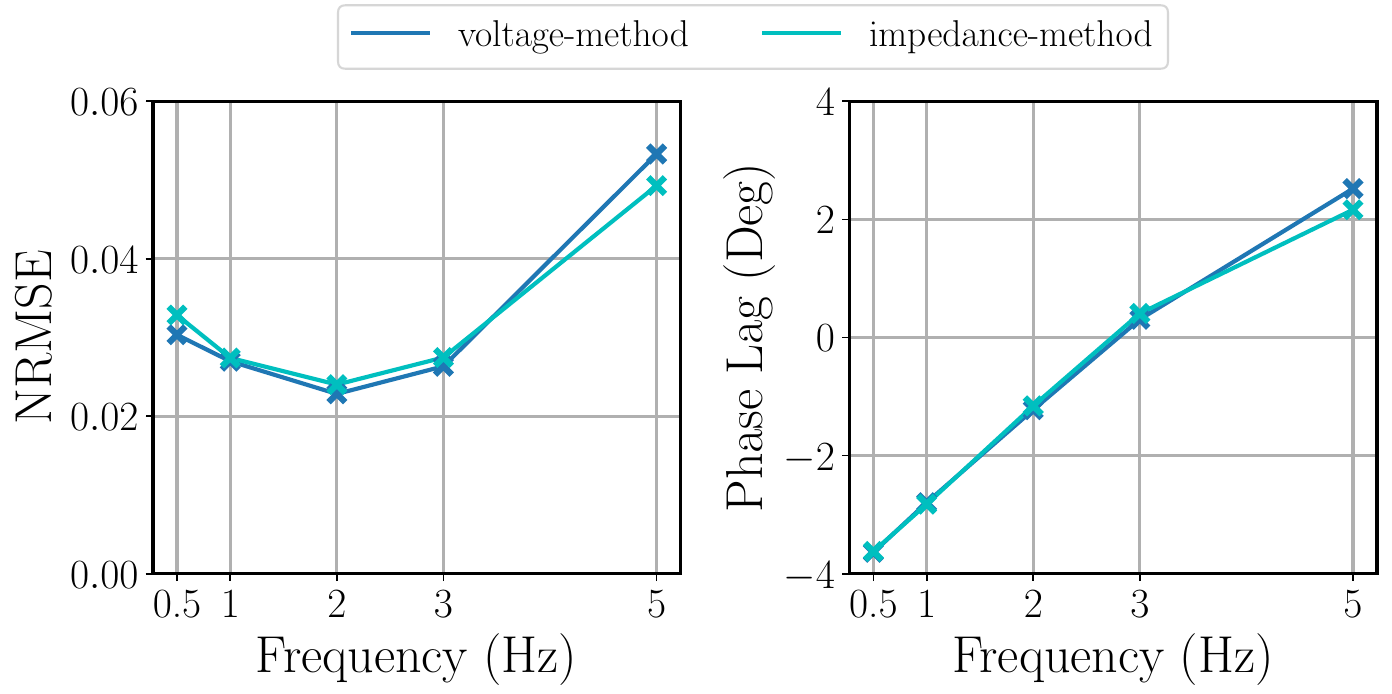}
   \caption{NRMSE and phase lag of the voltage-method and impedance-method respectively as functions of sinusoidal actuation frequencies at \SI{1}{kV} amplitude and \SI{3.5}{kV} offset.}
   \label{fig:plot2}
\end{figure}

Previous work in capacitive sensing for Peano-HASEL actuators has shown limitations when used together with miniature HV power supplies \cite{ly2021miniaturized}. This is due to a distortion of the calculated capacity for HV signals above 3.3 kV. The authors believe the cause to be a larger interfering noise occurring at the same frequency but out of phase with the sensing signal.
 
 To validate if the methods proposed in this paper enable the use of miniature power supplies past what was possible in previous work, we observe three different differential voltages, (\(V_H\)), (\(V_C\)), and (\(V_K\)), during a constant actuation of the F-HASEL at a driving voltage (\(V_{DC}\)) of \SI{4.8}{kV} (Fig. \ref{fig:4}). The voltage (\(V_H\)), measured across the LV sensing electrodes on the F-HASEL, as well as the voltage (\(V_C\)), measured across the measuring resistor (\(R_C\)), are analogous to the measurements captured by our proposed capacitive sensing circuitry. A \SI{10}{kOhm} measuring resistor (\(R_K\)), placed in the high-voltage signal path between the high-voltage electrode at low voltage and ground, is analogous in both placement and resistance to the measuring resistor used in a previously proposed method by Ly et al. \cite{ly2021miniaturized}. As all methods rely on measuring changes in an LV sinusoidal signal across the respective voltage measurements, we can qualitatively observe the difference in signal-to-noise ratio. To only observe the noise present in the measurements, no LV sinusoidal signals were applied.

As is evident from Fig. \ref{fig:4}, measurements (\(V_K\)) and (\(V_C\)) are in the same order of magnitude, with a calculated RMS of \SI{618.6}{mV} for (\(V_K\)) and \SI{1.382}{V} for (\(V_C\)). This is of note as the resistances of (\(R_K\)) and (\(R_C\)) differ by two orders of magnitude, with (\(R_K\)) having a resistance of \SI{10}{kOhm} and (\(R_C\)) a resistance of \SI{1}{MOhm}. A likely explanation for how measurements (\(V_K\)) and (\(V_C\)) can be in the same order of magnitude despite the resistance differing by a large amount lies in the mechanism through which the noise presents itself on the respective measurements. Measuring resistor (\(R_K\)) is placed in the direct signal path of all noise generated by the HV power supply. Measuring resistor (\(R_C\)) however is only affected by noise which is coupled onto the respective sensing circuitry through parasitic capacitive coupling between the HV electrodes and the LV sensing electrodes. 

The main takeaway from Fig. \ref{fig:4} however, is the reduction of noise on the voltage measurement (\(V_H\)) across the LV sensing electrodes compared to (\(V_K\)) and (\(V_C\)). With a calculated RMS value of \SI{37.19}{mV}, this equals a reduction in noise of a factor of around \SI{17}{} compared to (\(V_K\)). We believe that this is due to the additional resistor (\(R_N\)) placed on the high-side of the LV sensing electrodes, identical in resistance to the measurement resistor (\(R_C\)). Due to the symmetry of the setup, the noise coupled to the sensing circuitry presents itself as a common mode signal across the LV sensing electrodes. As the instrumentation amplifier used to measure the voltage across the LV sensing electrodes only measures the differential voltage, this common mode signal is rejected.

The reduction in noise across the LV sensing electrodes, compared to the resistors (\(R_K\)) and (\(R_C\)), means that any sensing method based on measuring changes in the voltage (\(V_H\)) will have a notably higher signal-to-noise ratio than methods measuring changes across measuring resistors (\(R_C\)) or (\(R_K\)), assuming the same LV signal is applied. While the increase in signal-to-noise ratio can be assumed to be around an order of magnitude, it is of note that this is only a qualitative estimation. The absolute magnitude of a LV sinusoidal signal across (\(V_H\)), (\(V_K\)), and (\(V_C\)) respectively depends on the capacitances between both the HV electrodes and the LV sensing electrodes. These capacitances are not constant but change during the actuation of the actuator. Furthermore, while the noise on (\(V_K\)) is a direct result of the measuring resistor (\(R_K\)) being placed in the signal path of the HV driving signal, the noise on (\(V_H\)) and (\(V_C\)) is a result of parasitic coupling which is susceptible to changes in the experimental setup, \textit{e.g.}, the placement of signal cables or the isolation of components. Additionally, as the reduction of noise on (\(V_H\)) is at least partially attributed to the common mode rejection ratio of the instrumentation amplifier used in the setup, the performance of the component itself affects the quantitative ratio.

\subsection{Setup}\label{AA}

The setup for our experiments involved mounting the HASEL vertically on a custom rig, as shown in Fig. \ref{fig:setup}A. The top of the HASEL was securely fixed on the rig, while an interchangeable weight was attached to the bottom of the HASEL. Unless noted otherwise, all experiments were conducted with a total weight of \SI{47.8}{g} attached.

A custom MATLAB script generates the actuation input signal and transmits it through a DAQ (USB-6343, National Instruments) to a miniature HV power supply (HVA0560, HVM Technology Inc.). The HV power supply amplifies the received input signal between \SI{0}{} to \SI{5}{V} to a high-voltage output signal in the range of \SI{0}{} to \SI{6}{kV}. This high-voltage output signal is connected to the HV electrodes on our HASEL actuator, enabling controlled actuation during our experiments. 

The high-frequency LV sensing signal (\SI{2}{kHz} sine wave) is generated by a signal generator (SFG-1013; GW Instek) and connected to the input of the sensing circuit. The two voltage measurements generated by the sensing circuit are recorded by the DAQ and read at a rate of \SI{100}{kHz} by the MATLAB script. For each sine wave of the two voltage measurements received, the average RMS over \SI{200}{} samples is calculated by the MATLAB script. The ground truth for the displacement was recorded by a laser point sensor (OM70, Baumer) and read by the MATLAB script through the DAQ. 

\begin{figure}[t]
\centering
\includegraphics[width=0.49\textwidth]{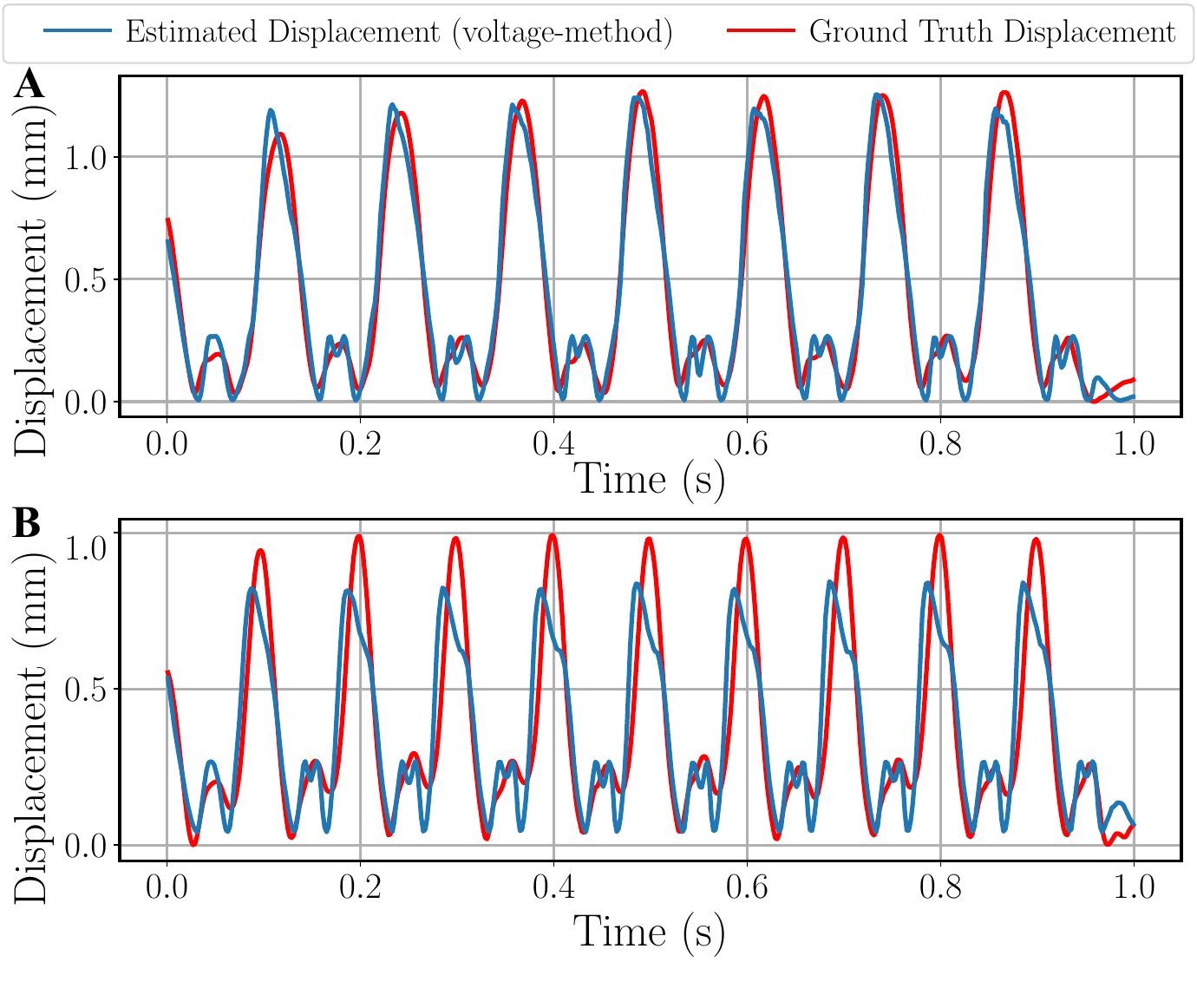}
   \caption{Displacement estimation using the voltage-method of a sinusoidal actuation with amplitude \SI{1}{kV} and offset \SI{3.5}{kV} at (A) \SI{8}{Hz} and (B) \SI{10}{Hz}.}
   \label{fig:plot3}
\end{figure}

\subsection{Displacement estimation during high-frequency actuation of the actuator}

In this section, we assess the performance of our capacitive sensing method for actuation signals at frequencies between \SI{0.5}{Hz} and \SI{20}{Hz}. It is worth noting that no analog or digital filter was applied to the measurements unless otherwise indicated. Employing an offline third-order polynomial mapping approach, we mapped the results of both (1) the voltage-method and (2) the impedance-method to the ground truth displacement recorded by the high-precision laser for each actuation cycle.

As shown in Fig.~\ref{fig:plot1}, the estimated displacement generated by using the voltage-method shows a close overlap with the ground truth displacement for both a step actuation signal at \SI{0.5}{Hz} as well a sinusoidal actuation signal at frequencies from \SI{0.5}{Hz} up to \SI{5}{Hz}. In Fig.~\ref{fig:plot2}, we visualize the normalized root mean squared error (NRMSE) and the phase lag between the displacement estimation and ground truth displacement for both the voltage-method and the impedance-method. From this graph, it becomes clear that the proposed method shows a drastic improvement in accuracy at estimating actuator displacement at frequencies of \SI{1}{Hz} and above compared to previous methods. Compared to the work of Ly et al. \cite{ly2021miniaturized}, which this paper considers the state-of-art in capacitive sensing of Peano-HASEL actuators, we see at \SI{1}{Hz} a reduction from an NRMSE of approx. \SI{0.1000}{} to \SI{0.0302}{}, and at \SI{5}{Hz} a reduction from an NRMSE of approx. \SI{0.8000}{} to \SI{0.0524}{}. Additionally, the results in Fig.~\ref{fig:plot2} were all taken during actuation with a miniature power supply, unlike in the work of Ly et al. \cite{ly2021miniaturized}. This increase in performance at higher frequencies is largely attributable to the reduction in phase lag between the estimation and the ground truth, with the phase lag staying under \SI{4}{} degrees in either direction. This is a drastic reduction compared to a phase lag of approx. \SI{-32.3}{} degrees at an actuation frequency of \SI{5}{Hz} demonstrated in the work by Ly et al. \cite{ly2021miniaturized}.

Comparing the performance of the voltage-method to the one of impedance-method (Fig. \ref{fig:plot2}), we observe a trend of the voltage-method performing better at a frequency of \SI{3}{Hz} and below. Meanwhile the impedance-method performs better at higher frequencies, a trend which continues at frequencies of \SI{8}{Hz} and \SI{10}{Hz}, with the impedance-method showing a NRMSE of \SI{0.0589}{} at \SI{8}{Hz} and \SI{0.1414}{} at \SI{10}{Hz}, compared to the voltage-method with a NRMSE of \SI{0.0638}{} at \SI{8}{Hz} and \SI{0.1467}{} at \SI{10}{Hz}. We believe the cause to be that the performance of the different methods is affected to a different degree by different sources of noise, \textit{e.g.}, HV power supply ripple, or input bias current of the amplifiers. This results in the voltage-method and the impedance-method being more suitable for different applications.

\begin{figure}[t]
\centering
\includegraphics[width=0.5\textwidth]{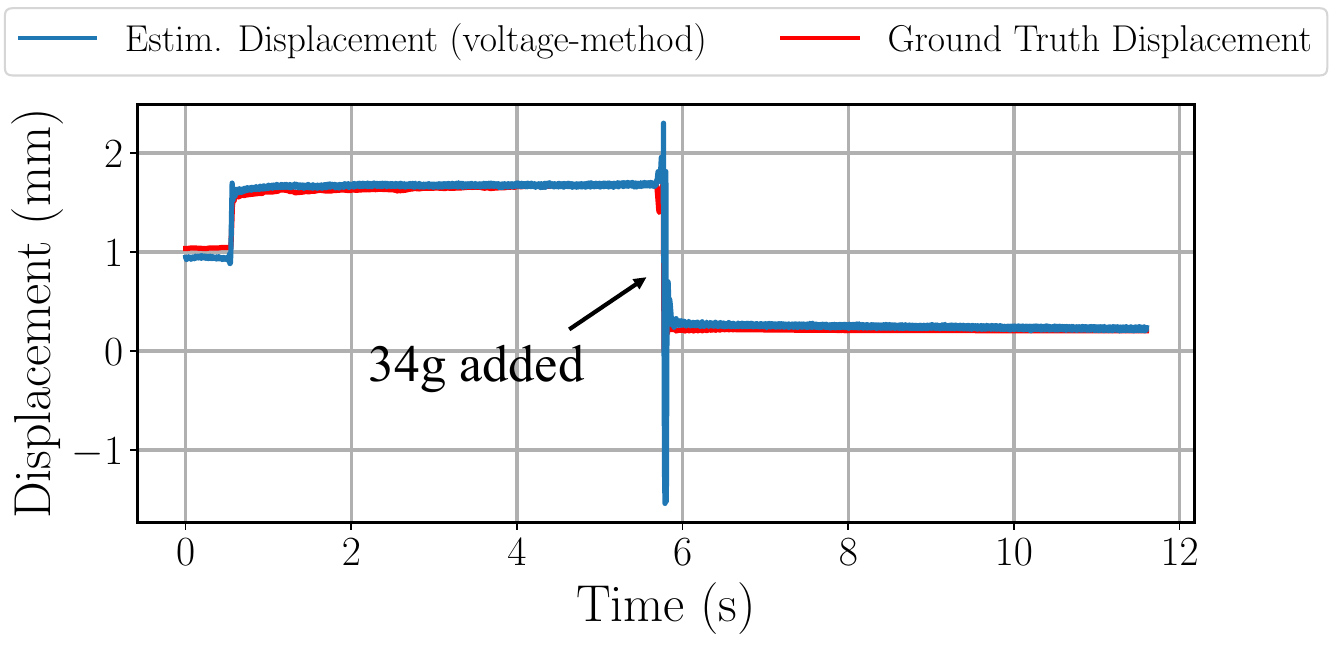}
   \caption{Displacement estimation using the voltage-method during a constant actuation of \SI{4}{kV} amplitude applied at t=\SI{0.5}{s} and an additional load of \SI{34}{g} added to the actuator at approx. t=\SI{6}{s}}.
   \label{fig:passive}
\end{figure}

\begin{figure*}[t]
\centering
\includegraphics[width=\textwidth]{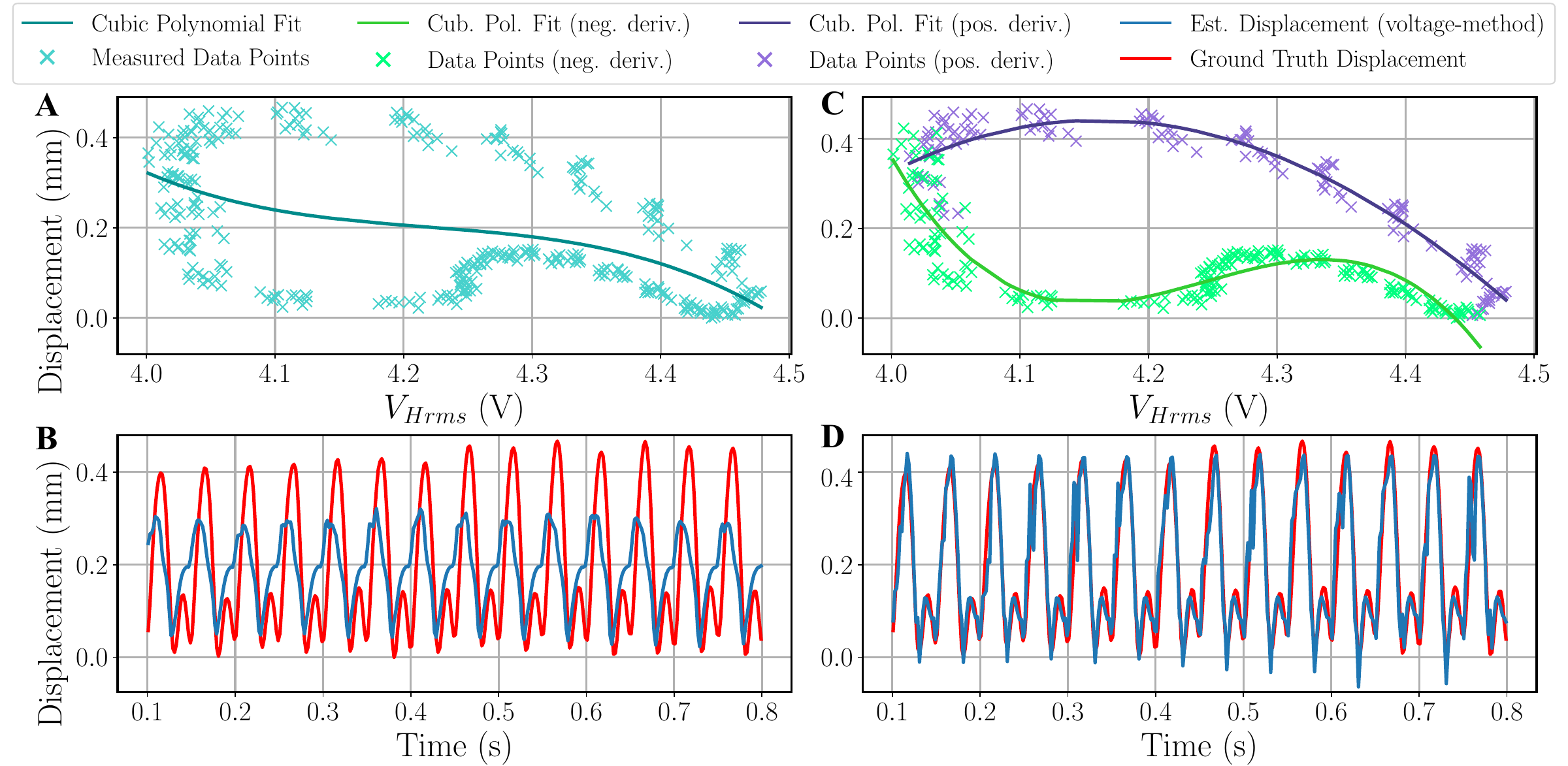}
   \caption{For a sinusoidal actuation frequency of \SI{20}{Hz} with \SI{1}{kV} amplitude and \SI{3.5}{kV} offset. (A) Third-order polynomial mapping approach. (B) Dual third-order polynomial mapping approach to enable accurate displacement estimation despite hysteresis effect. (C) Displacement estimation using the voltage-method and a third-order polynomial mapping approach. (D) Displacement estimation using the voltage-method and a dual third-order polynomial mapping approach.}
   \label{fig:polfit}
\end{figure*}

\subsection{Hysteresis of ground truth displacement as a function of voltage measurements}

As shown in Fig. \ref{fig:plot3}A, the estimated displacement generated by using the voltage-method shows good overlap with the ground truth displacement even at frequencies of \SI{8}{Hz}, despite what are assumed to be charge retention effects are starting to influence the actuator displacement at the lower peaks of the sinusoidal displacement. However, there is a drastic loss in performance when increasing the actuation frequency from \SI{8}{Hz} to \SI{10}{Hz} (Fig. \ref{fig:plot3}B). This trend of a drastic loss in performance continues as we increase the actuation frequency to \SI{20}{Hz} (Fig. \ref{fig:polfit}B). The cause for this loss in performance can be found in the fitting plot for an actuation frequency of \SI{20}{Hz} shown in Fig. \ref{fig:polfit}A. We observe a clearly defined hysteresis effect of the ground truth displacement as a function of the voltage (\(V_{Hrms}\)) measured across the LV sensing electrodes. Due to this hysteresis effect, our third-order polynomial mapping approach starts to fail at frequencies of around \SI{10}{Hz}. To address this, we propose a dual third-order polynomial mapping approach, which considers whether the actuator is contracting or relaxing at the time the displacement estimation takes place (Fig.~\ref{fig:polfit}C). This is implemented by using past voltage values to calculate the slope of the current displacement trajectory. As shown in Fig.~\ref{fig:polfit}D, this drastically improves sensing performance, bringing the NRMSE at a \SI{20}{Hz} actuation frequency from \SI{0.2615}{} down to \SI{0.1002}{}.

\subsection{Displacement estimation during a change in applied external tensile forces}

In this section, we assess the performance of the displacement estimation when external tensile forces cause a displacement of the F-HASEL despite a constant driving voltage of \SI{4.5}{kV} being applied to the HV electrodes. Fig. \ref{fig:passive} shows a relaxed F-HASEL being actuated with a constant driving voltage and an additional weight of \SI{34}{g} being added to the load attached to the actuator at a specific point in the actuation cycle. With a starting weight of \SI{13.8}{g}, the addition of \SI{34}{g} to the load increases the external tensile force from \SI{0.135}{N} to \SI{0.468}{N}. This increase in external tensile force causes a step in actuator displacement as the actuator is eccentrically contracted. We observe the displacement estimation being able to follow both the step in displacement caused by the application of a driving voltage and the step in displacement caused by the increase in the external tensile force acting on the actuator. The only considerable mismatch between the estimated and ground truth displacement occurs through a substantial overshoot of the displacement estimation when the external force is increased.

\subsection{Enabling multiple joint tracking in a VR application}

\begin{figure}[t]
\centering
\includegraphics[width=0.5\textwidth]{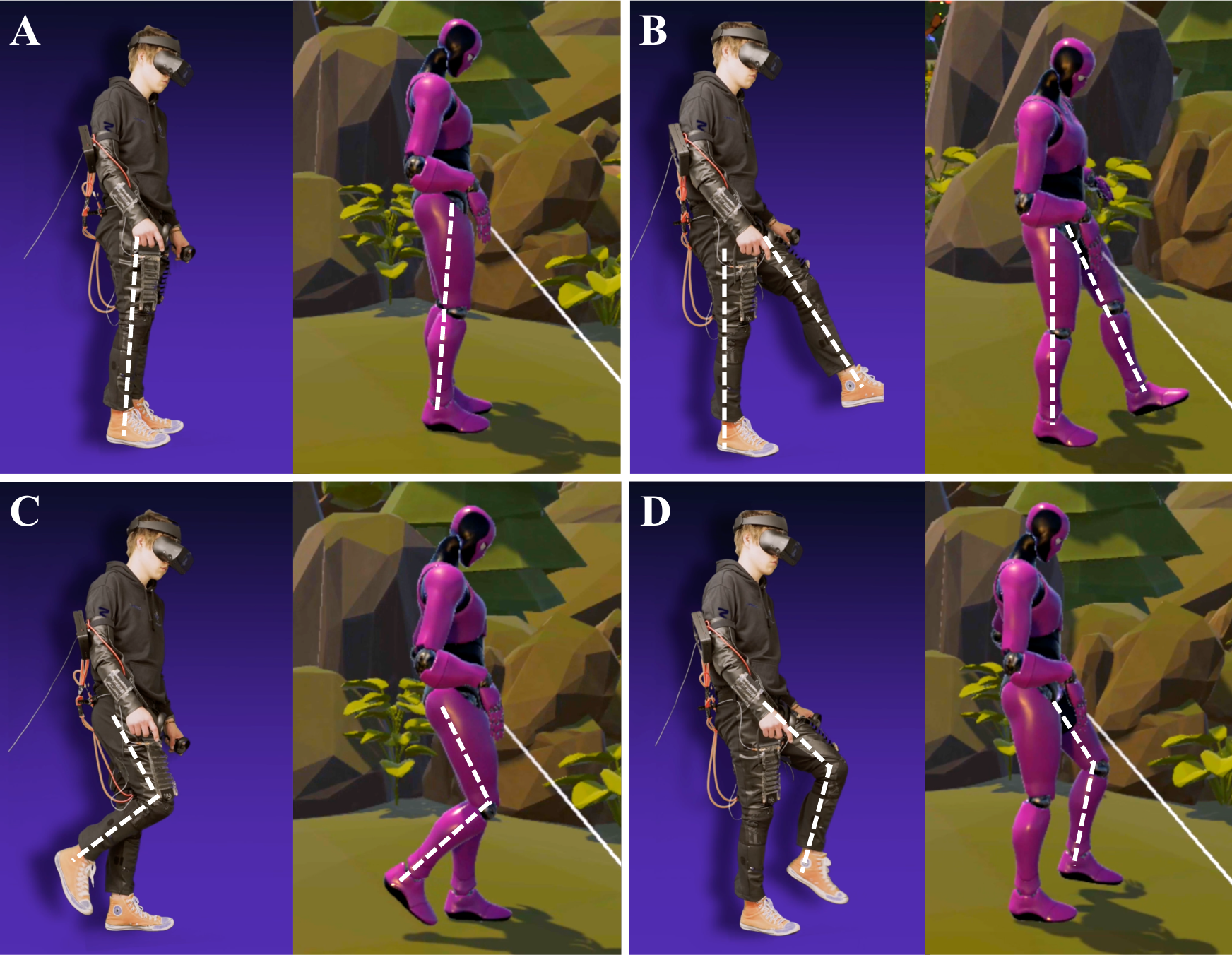}
   \caption{Real-time tracking of a VR user's hip and knee joint rotations in a 2D-plane with simultaneous transmission to the joints of a virtual avatar. The figure shows (A) a neutral upright pose, (B) hip flexion, (C) knee flexion, and (D) knee and hip flexion.}
   \label{fig:rig}
\end{figure}

In this section, we will use the proposed circuitry for quasi-simultaneous displacement estimation of multiple F-HASEL actuators to build a wearable application to track the joint rotation angles of the knee and the hip joint of a VR user in a 2D-plane (Fig. \ref{fig:1}A) and transmit them to the user's virtual avatar (Fig. \ref{fig:1}D). The following characteristics of our proposed sensing circuitry are what allows this application to be portable and untethered: (1) the ability to use a single compact sensing circuitry to measure multiple actuators; (2) the ability to ensure accurate sensing performance even when using a miniature HV power supply; (3) the demonstrated ability to measure changes in actuator displacement caused by external tensile forces. Four multi-pouch HASEL actuators are used, each consisting of 5 Peano-HASEL pouches in series with an F-HASEL pouch. All actuators are mounted on one end at a fixed point on a pair of trousers made from electrically isolating material. To transmit joint movement to an actuator displacement, the free end of the actuators are each connected to an inelastic string, which is guided over the respective joint to another fixed point on the trousers. The main purpose of the Peano-HASEL pouches is to provide the strain necessary for when the path between the fixed points over the joint increases as the joint rotates. This is analogous to the muscle-tendon system that enables the rotation of joints through muscles in the human body. The main difference is that in this application, none of the HASEL actuators assert a high enough momentum on the joints to cause any rotation. Instead, all HASEL pouches are actuated at a constant driving voltage of \SI{4}{kV}, and the application relies on the contracted HASEL pouches being pulled apart by an external tensile force transmitted through the inelastic strings as the respective joints are flexed. This results in a displacement of all pouches, including the F-HASEL, estimated by the sensing circuitry. During an offline calibration, the estimated actuator displacement is mapped to a webcam's ground truth joint rotations. After the full flexion of each tracked joint has been completed at least once during the calibration, the application can simultaneously estimate the rotations of four separate joints in real time and transmit the rotations to the joints of a virtual avatar in a unity program (Fig.~\ref{fig:rig}A-D).

\section{Conclusion}\label{conclusion}
In this paper, we introduce the F-HASEL as well as miniature low-voltage capacitive sensing circuitries, which take advantage of an electrode pair on the actuator used exclusively for sensing to enable accurate displacement estimations during high-frequency actuation with miniature power supplies. Additionally, the proposed sensing methods are all agnostic to the polarity of the driving voltage, allowing the mitigation of charge retention effects occurring in the actuator. We validate the displacement estimation performance of the proposed sensing methods for actuation frequencies up to \SI{20}{Hz}, as well as for changes in the external tensile force acting on the actuator, all of which while actuating with a miniature power supply. We demonstrate a wearable application in which we track the rotations of multiple joints of a VR user in a 2D-plane in real-time, using a sensing circuitry that can sense multiple F-HASELs quasi-simultaneously with minimal additional components. All of this combined demonstrates the potential of the proposed sensing methods to facilitate low-cost, portable, and untethered applications that make use of multiple soft actuators operating at high frequencies.

In future work, the LV sensing electrode pair's geometry, size, and position can be optimized to maximize sensing performance. Furthermore, as a large portion of the interfering noise present on measurements in the circuitry occurs through parasitic coupling, the sensing performance might be improved through the placement and isolation of different components in the circuitry.

%%%%%%%%%%%%%%%%%%%%%%%%%%%%%%%%%%%%%%%%%%%%%%%%%%%%%%%%%%%%%%%%%%%%%%%%%%%%%%%%

%%%%%%%%%%%%%%%%%%%%%%%%%%%%%%%%%%%%%%%%%%%%%%%%%%%%%%%%%%%%%%%%%%%%%%%%%%%%%%%%

%%%%%%%%%%%%%%%%%%%%%%%%%%%%%%%%%%%%%%%%%%%%%%%%%%%%%%%%%%%%%%%%%%%%%%%%%%%%%%%%
\iffalse
\section*{APPENDIX}

\begin{itemize}
    \item Further resources on mono, stereo, and event camera latency.
    \item Datasheet extracts will be used to back up our claims, if needed.
    \item Elaboration of the testing setup.
\end{itemize}
\fi

%%%%%%%%%%%%%%%%%%%%%%%%%%%%%%%%%%%%%%%%%%%%%%%%%%%%%%%%%%%%%%%%%%%%%%%%%%%%%%%%

% Bibliography - max 20! citations
\bibliographystyle{IEEEtran}

\bibliography{references}

\end{document}